\documentclass[11pt]{article}
\usepackage{acl2016}
\usepackage{times}
\usepackage{latexsym}
\usepackage{url}
\usepackage{caption}
\usepackage{subcaption}
\usepackage{tabularx}
\usepackage{booktabs}
\usepackage{multirow}
\usepackage{graphicx}
\usepackage{amsmath}
\usepackage{comment}
\usepackage{xcolor,colortbl}
\usepackage{enumitem}

\definecolor{light-gray}{gray}{0.9}
\definecolor{dark-gray}{gray}{0.6}
\definecolor{mid-gray}{gray}{0.7}

\aclfinalcopy 

\title{Learning Fine-Grained Knowledge about Contingent Relations \\ between Everyday Events}

\author{Elahe Rahimtoroghi, Ernesto Hernandez \and Marilyn A Walker \\
 Natural Language and Dialogue Systems Lab \\ Department of Computer Science, University of California Santa Cruz \\Santa Cruz, CA 95064, USA \\ {\tt elahe@soe.ucsc.edu, eherna23@ucsc.edu, mawalker@ucsc.edu}}
\date{}

\begin{document}

\maketitle

\begin{abstract}
Much of the user-generated content on social media is provided by
ordinary people telling stories about their daily lives.  We develop
and test a novel method for learning fine-grained common-sense
knowledge from these stories about contingent (causal and conditional)
relationships between everyday events.  This type of knowledge is
useful for text and story understanding, information extraction,
question answering, and text summarization. We test and compare
different methods for learning contingency relation, and compare what is learned
from topic-sorted story collections vs. general-domain stories. Our experiments show that using topic-specific datasets enables learning finer-grained knowledge about events and results in significant improvement over the baselines.  
An evaluation on Amazon Mechanical Turk shows 82\% of the 
relations between events that we learn from topic-sorted stories are
judged as contingent. 

\end{abstract}

\section{Introduction} 
\label{sec:intro}

The original idea behind scripts as introduced by Schank was to capture knowledge
about the fine-grained events of everyday experience, such as
\textit{opening a fridge} enabling \textit{preparing food}, or the
event of \textit{getting out of bed} being triggered by \textit{an
  alarm going off} \cite{Schank77}. This idea has
motivated previous work exploring whether common-sense knowledge about
events can be learned from text, however, only a few learn from data
other than newswire \cite{Huetal13,Manshadietal08,BeamerGirju09}.
News articles (obviously) cover newsworthy topics such as
\textit{bombing, explosions, war} and \textit{killing} so the
knowledge learned is limited to those types of events.

\begin{figure}[t!]
\small
\begin{tabularx}{\columnwidth}{X}
\toprule
\bf Camping Trip \\
\midrule
{\bf We packed all our things} on the night before Thu (24 Jul) except for frozen food. We brought a lot of things along. {\bf We woke up} early on Thu and JS started packing the frozen marinatinated food inside the small cooler... In the end, we decided the best place to set up the tent was the squarish ground that's located on the right. Prior to setting up our tent, {\bf we placed a tarp on the ground}. In this way, the underneaths of the tent would be kept clean. After that, {\bf we set the tent up}.  \\
\toprule
\bf Storm \\
\midrule

I don't know if I would've been as calm as I was without the radio, as
 {\bf the hurricane made landfall} in Galveston at 2:10AM on Saturday. As
{\bf the wind blew}, branches thudded on the roof or trees snapped, it
was helpful to pinpoint the place...  {\bf A tree fell} on the garage
roof, 
but it's minor damage compared
to what could've happened. We then {\bf started cleaning up}, despite
Sugar Land implementing a curfew until 2pm; I didn't see any policemen
enforcing this. Luckily my dad has a gas saw (as opposed to electric),
so {\bf we helped cut up} three of our neighbors' trees. {\bf I did a
  lot of raking}, and there's so much debris in the
garbage. 
\\ \bottomrule
\end{tabularx}
\caption{Excerpts of two stories in the blogs corpus on the topics
of {\it Camping Trip} and {\it Storm}.}
\label{fig:story-example}
\end{figure}

However, much of the user-generated content on social media is provided
by ordinary people telling stories about their daily lives. These
stories are rich with common-sense knowledge. For example, the {\it
  Camping Trip} story in Fig.~\ref{fig:story-example} contains
implicit common-sense knowledge about contingent (causal and
conditional) relations between camping-related events, such as {\it
  setting up a tent} and {\it placing a tarp}. The {\it Storm} story
contains implicit knowledge about events such as {\it the hurricane
  made landfall}, {\it the wind blew}, {\it a tree fell}.  Our aim is
to learn fine-grained common-sense knowledge about contingent
relations between everyday events from such stories.  We show that the
fine-grained knowledge we learn is simply not
found in publicly available narrative and event schema
collections~\cite{ChambersJurafsky09,Balasubramanianetal13}.

Personal stories provide both advantages and disadvantages for
learning common-sense knowledge about events. An advantage is that
they tend to be told in chronological order~\cite{GordonSwanson09},
and temporal order between events is a strong cue to contingency
\cite{Prasadetal08,BeamerGirju09}. However, their structure is more
similar to oral narrative than to newswire
\cite{Rahimtoroghietal14,Swansonetal14b}. Only about a third of
the sentences in a personal narrative describe actions,\footnote{The other two
  thirds provide scene descriptions and descriptions of the thoughts
  or feelings of the narrator.} so novel methods are needed
to find useful relationships between events.

Another difference between our work and prior research is that much of the work on narrative
  schemas, scripts, or event schemas characterize what is
learned as ``collections of events that tend to co-occur''. Thus what
is learned is not evaluated for contingency
~\cite{ChambersJurafsky08,ChambersJurafsky09,Manshadietal08,Nguyenetal15,Balasubramanianetal13,PichottaMooney14}. 
Historically, work on scripts explicitly modeled causality
\cite{Lehnert81} {\it inter alia}.
Our work is motivated by Penn Discourse Treebank (PDTB) definition of {\sc contingency} that has two types: {\sc cause} and {\sc condition}, and is more similar to approaches that learn specific event
relations such as contingency or causality
\cite{Huetal13,DoChaRo11,Girju03,RiazGirju10,Rinketal10,ChklovskiPantel04}. 
Our contributions
are as follows:
\begin{itemize}
\item We use a corpus of everyday events for learning common-sense knowledge focusing on the contingency relation between events. We first use a subset of the corpus including general-domain stories. Next, we produce a 
topic-sorted set of stories using a semi-supervised bootstrapping method to learn finer-grained knowledge. We
use two different datasets to directly compare what is learned from topic-sorted stories as opposed to a general-domain story corpus (Sec.~\ref{sec:data});

\item We develop a new method for learning contingency relations
between events that is tailored to the ``oral narrative'' nature
of blog stories. We apply Causal Potential~\cite{BeamerGirju09} to model the contingency relation between two events. We directly compare our method to several other approaches as baselines (Sec.~\ref{sec:events}).
We also identify topic-indicative contingent event pairs from our topic-specific corpus that can be used as building blocks for generating coherent event chains and narrative schema for a particular theme (Sec.~\ref{sub:indic});

\item We conduct several experiments to evaluate the quality
  of the event knowledge learned in our work that indicate our results are contingent and topic-related. We directly compare the common-sense knowledge we learn with the Rel-grams collection and show that what we learn is not found in available corpora (Sec.~\ref{sec:eval}).
\end{itemize}

We release our contingent event pair collections for each topic for future use of other research groups~\footnote{https://nlds.soe.ucsc.edu/everyday\_events}.

\section{A Corpus of Everyday Events} 
\label{sec:data} 

Our dataset is drawn from the Spinn3r corpus of
millions of blog
posts~\cite{Burtonetal09,GordonSwanson09,Gordonetal12}.  
We hypothesize that personal stories are a valuable resource to
learn common-sense knowledge about relations between
everyday events and that finer-grained knowledge can be learned from
topic-sorted stories~\cite{RiazGirju10} that share a particular theme, so we construct two different
sets of stories: 

\noindent \textbf{General-Domain Set.} We created a random subset from the Spinn3r corpus from personal blog domains: \emph{livejournal.com, wordpress.com, blogspot.com, spaces.live.com, typepad.com, travelpod.com}. This set consists of 4,200 stories not selected for any specific topic.

\noindent \textbf{Topic-Specific Set.} We produced a dataset by filtering the corpus using a bootstrapping method to create topic-specific sets for topics such as {\it going camping, being arrested, going snorkeling or scuba diving, visiting the dentist, witnessing a
  major storm}, and {\it holiday activities} associated with
Thanksgiving and Christmas (see Table~\ref{tab:topics}).

\begin{table}[t]
\footnotesize
\centering
\begin{tabularx}{\columnwidth}{p{0.6in} | X}
\toprule
\bf Topic & \bf Events \\
\midrule
Camping Trip & camp(), roast(dobj:marshmallow), hike(), pack(), fish(), go(dobj:camp), grill(), put(dobj:tent , prt:up), build(dobj:fire) \\
\midrule
Storm & restore(), lose(dobj:power), rescue(), evacuate(), flood(), damage(), sustain(), survive(), watch(dobj:storm) \\
\midrule
Christmas Holidays & open(dobj:present), exchange(dobj:gift), wrap(), sing(), play(), snow(), buy(), decorate(dobj:tree), celebrate() \\
\midrule
Snorkeling and Scuba Diving &  see(dobj:fish), swim(), snorkel(), sail(), surface(), dive(), dart(), rent(dobj:equipment), enter(dobj:water), see(dobj:turtle) \\
\bottomrule
\end{tabularx}
\caption{\label{tab:topics} Some topics and examples of their indicative events.}
\end{table}

We apply AutoSlog-TS, a semi-supervised algorithm that learns
narrative \textit{event-patterns} to bootstrap a collection of stories
on the same theme \cite{Riloff96}.  These patterns, developed for
information extraction, search for the syntactic constituent with the
designated word as its head. For example, consider the example in the
first row of Table~\ref{tab:CF-example}: {\small{\tt
    NP-Prep-(NP):CAMPING-IN}}. This pattern looks for a \textit{Noun
  Phrase (NP)} followed by a \textit{Preposition (Prep)} where the
head of the NP is {\sc camping} and the Prep is {\sc in}.  Our
algorithm consists of the following steps for each topic: \\

\noindent \textbf{1.~~Hand-labeling:} 
We manually labeled a small set ($\sim$ 200-300) of stories on the topic.  

\noindent \textbf{2.~~Generating Event-Patterns:} 
Given hand-labeled stories on a topic (from Step 1), and
a random set of stories that are not relevant to that topic, AutoSlog-TS learns a set of syntactic templates (case frame templates) that distinguish the
linguistic patterns characteristic of the topic from the random
set. For each pattern it generates frequency and conditional probability which indicate how strongly the pattern is associated with the topic.

Table~\ref{tab:CF-example} shows examples of such patterns that we have learned for two different topics. We call them \textit{indicative event-patterns} for each topic. Table~\ref{tab:topics} shows examples of the indicative event-patterns for different topics. They are mapped to our event representation described in Sec~\ref{sec:events}, e.g., the pattern {\small{\tt (subj)-ActVB-Dobj:WENT-CAMPING}} in
Table~\ref{tab:CF-example} is mapped to {\small{\tt go(dobj:camp)}}.



\noindent \textbf{3.~~Parameter Tuning:}
We use the frequency and probability generated by AutoSlog-TS and apply a
threshold for filtering to select a subset of indicative
event-patterns strongly associated with the topic. 
In this step we aim to find optimal values
for frequency and probability thresholds denoted as
\textit{f-threshold} and \textit{p-threshold} respectively.  We
divided the hand-labeled data from Step 1 into train and development
sets and designed a classifier based on our bootstrapping method:
if the number of event-patterns extracted from a post is more than a
certain number (\textit{n-threshold}), it is labeled as positive and otherwise it is labeled as negative meaning that
it is not related to the topic.  We repeated the classification for
several combinations of different values for each of the three
parameters and measured the precision, recall and f-measure. We
selected the optimal values for the thresholds that resulted in high
precision (above 0.9) and average recall (around 0.4). We compromised
on a lower recall to achieve a high precision to establish a highly
accurate bootstrapping algorithm. Since bootstrapping is performed
on a large set of stories, a low recall stills result in
identifying enough stories per topic. 

\begin{table}[t]
\centering
\small
\begin{tabularx}{\columnwidth}{p{0.75in} | X}
\toprule
{\bf Topic} & {\bf Event-Pattern (Case Frame) Examples} \\
\midrule
Camping Trip & NP-Prep-(NP):CAMPING-IN \\ 
 & NP-Prep-(NP):HIKE-TO \\ 
 & (subj)-ActVB-Dobj:WENT-CAMPING \\
 & NP-Prep-(NP):TENT-IN \\
\midrule
Storm & (subj)-ActVp-Dobj:LOST-POWER \\ 
 & (subj)-ActVp:RESTORED \\ 
 & (subj)-AuxVp-Dobj:HAVE-DAMAGE \\
 & (subj)-ActVp:EVACUATED \\
\bottomrule
\end{tabularx}
\caption{Examples of narrative event-patterns (case frames) learned from corpus.}
\label{tab:CF-example}
\end{table}

\noindent \textbf{4.~~Bootstrapping:} We use the patterns learned in
previous steps as indicative event-patterns for the topic.  The
bootstrapping algorithm processes each story, using
AutoSlog-TS to extract lexico-syntactic patterns. Then it counts the indicative event-patterns in the extracted patterns,
and labels the blog as a positive instance for that topic if the 
count is above the n-threshold value for that topic.

The manually labeled dataset includes 361 Storm and 299 Camping Trip
stories. After one round of bootstrapping the algorithm identified
971 additional Storm and 870 more Camping Trip stories.  The
bootstrapping method is not evaluated separately, however, the results
in Sec.~\ref{sub:automatic-test} indicate that using the bootstrapped data
considerably improves the accuracy of the contingency model and enhances extracting topic-relevant event knowledge.

\section{Learning Contingency Relation between Narrative Events}
\label{sec:events}
In this section we describe our representation of events in narratives and our methods for modeling contingency relationship between events.

\subsection{Event Representation}
\label{sub:represent}
In previous work different representations have been proposed for the
event structure such as single verb and verb with two or more
arguments. Verbs are used as a central indication of an event in a narrative. However, other entities related to the verb also play
a strong role in conveying the meaning of the event. In~\cite{PichottaMooney14} it is shown that the
multi-argument representation is richer than the previous ones and is
capable of capturing interactions between multiple events.  
We use a
representation that incorporates the \textit{Particle} of the verb in
the event structure in addition to the \textit{Subject} and the
\textit{Direct Object} and define an event as a verb with its dependency
relations as follows:

\begin{quotation}
Verb Lemma (subj:Subject Lemma, dobj:Direct Object Lemma, prt:Particle) 
\end{quotation} 

Table~\ref{tab:rep} shows example sentences describing an event from
the Camping topic along with their event structure. The examples show
how including the arguments often change the meaning of an event. In
Row 1 the \textit{direct object} and \textit{particle} are required to
completely understand the event in this sentence. Row 2 shows another
example where the verb \textit{have} cannot implicate what event is
happening and the direct object \textit{oatmeal} is needed to
understand what has occurred in the story.

We parse each sentence and extract every verb lemma with its arguments
using Stanford dependencies~\cite{Manningetal14}. For each verb, we
extract the \textit{nsubj}, \textit{dobj}, and \textit{prt} dependency
relations if they exist, and use their lemma in the event
representation. To generalize the event
representations, we use the types identified by
Stanford's Named Entity Recognizer and map each argument to its named
entity type if available, e.g., in Row 3 of Table~\ref{tab:rep}, the
\textit{Lost Valley River Campground} is represented by its type {\sc
  location}. We use abstract types for named entities such as {\sc
  person}, {\sc organization}, {\sc time} and {\sc date}.  We also
represent each pronoun by the abstract type {\sc person}, e.g. Row 5
in Table~\ref{tab:rep}.

\begin{table}[t]
\footnotesize
\begin{tabularx}{\columnwidth}{p{0.1in} | X}
\toprule
\# & \bf Sentence $\rightarrow$ \bf Event Representation \\
\midrule
1 & but it wasn't at all frustrating \textit{putting up the tent} and setting up the first night $\rightarrow$ put (dobj:tent, prt:up) \\
\midrule
2 & The next day \textit{we had oatmeal} for breakfast \\ 
 & $\rightarrow$ have (subj:{\sc person}, dobj:oatmeal) \\
\midrule
3 & by the time \textit{we reached the Lost River Valley Campground}, it was already past 1 pm \\
 & $\rightarrow$ reach (subj:{\sc person}, dobj:{\sc location}) \\
\midrule
4 & then \textit{JS set up a shelter} above the picnic table \\
 & $\rightarrow$ set (subj:{\sc person}, dobj:shelter, prt:up) \\
\midrule
5 & once the rain stopped, \textit{we built a campfire} using the firewoods $\rightarrow$ build (subj:{\sc person}, dobj:campfire) \\
\bottomrule
\end{tabularx}
\caption{\label{tab:rep} Event representation examples from Camping Trip topic.}
\end{table}

\subsection{Causal Potential Method}
\label{sub:cp}

We define a \textit{contingent event pair} as a sequence of two events \textit{$(e_1,e_2)$} such that
$e_1$ and $e_2$ are likely to occur together in the given order and
$e_2$ is contingent upon $e_1$.  We apply an unsupervised
distributional measure called \textit{Causal Potential}  to induce the
contingency relation between two events.

Causal Potential (CP) was introduced
by~\newcite{BeamerGirju09} as a way to measure the tendency of an
event pair to encode a causal relation, where event pairs with high CP
have a higher probability of occurring in a causal context. 
We calculate CP for every pair of adjacent events in each topic-specific dataset. We
used a 2-skip bigram model which considers two events to be adjacent
if the second event occurs within two or less events after the first
one. 

We use skip-2 bigram in order to capture the
fact that two related events may often be separated by a non-essential
event, because of the oral-narrative nature of our data~\cite{Rahimtoroghietal14}. 
In contrast to the verbs that describe an event (e.g., \textit{hike, climb, evacuate, drive}), some verbs describe private states such as as \textit{belong, depend, feel, know}. 
We filter out clauses that tend to be associated with private states~\cite{Wiebe90}.
A pilot evaluation showed that this improves the results.
 
Equation~\ref{eq:cp} shows the formula for calculating Causal Potential of a pair consisting of two events: \textit{($e_1$, $e_2$)}. Here \textit{P} denotes probability and \textit{$P(e_1 \rightarrow e_2)$} is the probability of $e_2$ occurring after $e_1$ in the adjacency window which is equal to 3 due to the skip-2 bigram model. $P(e_2|e_1)$ is the conditional probability of $e_2$ given that $e_1$ has been seen in the adjacency window. This is equivalent to the Event-Bigram model described in Sec.~\ref{sub:baseline}.

\begin{equation}\label{eq:cp}
CP(e_1, e_2) = log\frac{P(e_2|e_1)}{P(e_2)} + log\frac{P(e_1 \rightarrow e_2)}{P(e_2 \rightarrow e_1)}
\end{equation}

To calculate CP, we need to compute event counts from the corpus and thus we need to define when two events are considered equal. The simplest approach is to define two events to be equal when their verb and arguments exactly match. However, with a close look at the data this approach does not seem adequate. For example, consider the following events:

\begin{quote}
go (subj:{\sc person}, dobj:camp) \\
go (subj:family, dobj:camp) \\
go (dobj:camp)
\end{quote}

\noindent They encode the same action although their representations
do not exactly match and differ in the subject. Our intuition is that
when we count the number of events represented as {\small{\tt go
    (subj:PERSON, dobj:camp)}} we should also include the count of
{\small{\tt go (dobj:camp)}}. To be able to generalize over the event
structure and take into account these nuances, we consider two events
to be equal if they have the same verb lemma and share at least one argument other than the subject.

\subsection{Baseline Methods}
\label{sub:baseline}
Our previous work on modeling contingency relations in film scripts data compared Causal Potential to methods used in previous work: Bigram event models \cite{Manshadietal08} and Pointwise Mutual
Information (PMI) \cite{ChambersJurafsky08} and the evaluations showed that CP obtains better results~\cite{Huetal13}.
In this work, we use CP for inducing contingency relation between events and apply three other models as baselines for comparison:

\noindent \textbf{Event-Unigram.} This method will produce a distribution of normalized frequencies for events.

\noindent \textbf{Event-Bigram.} We calculate the bigram probability of every pair of adjacent events using skip-2 bigram model using the Maximum Likelihood Estimation (MLE) from our datasets: 
\begin{equation}
P(e_2|e_1) = \frac{Count(e_1, e_2)}{Count(e_1)}
\end{equation}

\noindent \textbf{Event-SCP.} We use the Symmetric Conditional Probability between event tuples (Rel-grams) used in~\cite{Balasubramanianetal13} as another baseline method. The Rel-gram model is the most relevant previous work to our method and outperforms the previous state of the art on generating narrative event schema. This metric combines bigram probability considering both directions:
\begin{eqnarray}
SCP(e_1, e_2) = P(e_2|e_1) \times P(e_1|e_2)
\end{eqnarray}
Like Event-Bigram, we used MLE for estimating Event-SCP from the corpus.

\section{Evaluation Experiments}
\label{sec:eval}  

We conducted three sets of experiments to evaluate different aspects of our work. First, we compare the content of our topic-specific event pairs to current state of the art event collections to show that the fine-grained knowledge we learned about everyday events does not exist in previous work focused on the news genre. Second, we run an automatic evaluation test, modeled after the COPA task~\cite{Roemmele2011choice}, on a held-out test set to evaluate the event pair collections that we have extracted from both General-Domain and Topic-Specific datasets, in terms of contingency relations.
We hypothesize that the contingent event pairs can be used as basic elements for generating coherent event chains and narrative schema.
So, in the third part of the experiments, we extract topic-indicative contingent event pairs from our Topic-Specific dataset and run an experiment on Amazon Mechanical Turk (AMT) to evaluate the top N pairs with respect to their contingency relation and topic-relevance.

\begin{table}[t]
\centering
\small
\begin{tabularx}{\columnwidth}{ X | p{1in}}
\toprule
{\bf Label} & {\bf Rel-gram Tuples} \\
\midrule
Contingent \& Strongly Relevant & 7 \%  \\
Contingent \& Somewhat Relevant & 0 \%  \\
Contingent \& Not Relevant & 35 \%  \\
\midrule
Total Contingent & 42 \%  \\
\bottomrule
\end{tabularx}
\caption{Evaluation of Rel-gram tuples on AMT.}
\label{tab:relgram-result}
\end{table}

\subsection{Comparison to Rel-gram Tuple Collections}
\label{sub:relgram}
We chose Rel-gram tuples~\cite{Balasubramanianetal13} for comparison since it is the most relevant previous work to us: they generate pairs of relational tuples of events, called \textit{Rel-grams} using co-occurrence statistics based on Symmetric Conditional Probability described in Sec~\ref{sub:baseline}. 
Additionally, the Rel-grams are publicly available through an online search interface\footnote{http://relgrams.cs.washington.edu:10000/relgrams} and their evaluations show that their method outperforms the previous state of the art on generating narrative event schema.

However, their work is focused on news articles and does not consider the causal relation between events for inducing event schema.
We compare the content of what we learned from our topic-specific corpus to the Rel-gram tuples to show that the fine-grained type of knowledge that we learn is not found in their events collection. We also applied the co-occurrence statistics that they used on our data as a baseline (Event-SCP) for comparison to our method and present the results in Sec.~\ref{sub:automatic-test}.

In this experiment we compare the event pairs extracted from our
Camping Trip topic to the Rel-gram tuples.
The Rel-gram tuples are not sorted by topic. To find tuples relevant to Camping Trip, we used our top 10 indicative events and extracted all the Rel-gram tuples that included at least one event corresponding to one of the Camping Trip indicative events. For example, for {\small{\tt go(dobj:camp)}}, we pulled out all the tuples that included this event from the Rel-grams collection. The indicative events for each topic were automatically generated during the bootstrapping using AutoSlog-TS (Sec.~\ref{sec:data}). 

Then we applied the same sorting and filtering methods presented in
the Rel-grams work and removed any tuple with
frequency less than 25 and sorted the rest by the total symmetrical
conditional probability. These numbers are publicly available as
a part of the Rel-grams collection. We evaluated the top $N=100$ tuples of
this list using the Mechanical Turk task described later in
Sec.~\ref{sub:amt}. The evaluation results presented in
Table~\ref{tab:relgram-result} show that 42\% of the Rel-gram
pairs were labeled as contingent by the annotators and only 7\% were both contingent and topic-relevant. We argue that this
is mainly due to the limitations of the newswire data which does not
contain the fine-grained everyday events that we have extracted from our corpus.

\begin{table}[t]
\centering
\small
\begin{tabularx}{\columnwidth}{p{0.4in} X  p{0.35in}}
\toprule
{\bf Topic} & {\bf Dataset} & {\bf \# Docs} \\
\midrule
Camping & Hand-labeled held-out test & 107 \\
 Trip &  Hand-labeled train (Train-HL) & 192 \\
 & Train-HL + Bootstrap (Train-HL-BS) & 1,062 \\
\midrule
Storm & Hand-labeled held-out test & 98 \\
 &  Hand-labeled train (Train-HL) & 263 \\
 &  Train-HL + Bootstrap (Train-HL-BS) & 1,234 \\ 
\bottomrule
\end{tabularx}
\caption{Number of stories in the train and test sets from topic-specific dataset.}
\label{tab:train-test}
\end{table}

\subsection{Automatic Two-Choice Test}
\label{sub:automatic-test}

For evaluating our contingent event pair collections we have automatically generated a set of two-choice questions along with the answers, modeled after the COPA task~\cite{Roemmele2011choice}. 
We produced questions from held-out test sets for each dataset. Each question consists of one event and two choices. The \textit{question event} is one that occurs in the test data. One of the choices is an event adjacent to the question event in the document. The other choice is an event randomly selected from the list of all events occurring in the test set. The following is an example of a question from the Camping Trip test set:

\begin{quote}
{\bf Question event:} arrange (dobj:outdoor) \\
{\bf Choice 1:} help (dobj:trip) \\
{\bf Choice 2:} call (subj:{\sc person})
\end{quote}

\noindent In this example, {\small{\tt arrange (dobj:outdoor)}} is followed by the event {\small{\tt help (dobj:trip)}} in  a document from the test set and {\small{\tt call (subj:PERSON)}} was randomly generated. The model is supposed to predict which of the two choices is more likely to have a contingency relation with the event in the question. We argue that a strong contingency model should be able to choose the correct answer (the one that is adjacent to the question event) and the accuracy achieved on the test questions is an indication of the model's robustness.

\begin{table}[t]
\centering
\small
\begin{tabularx}{2.5in}{p{1.5in} p{1in}}
\toprule
{\bf Model} & {\bf Accuracy} \\
\midrule
Event-Unigram & 0.478 \\
Event-Bigram & 0.481 \\
Event-SCP (Rel-gram) & 0.477 \\
Causal Potential & 0.510 \\
\bottomrule
\end{tabularx}
\caption{Automatic two-choice test results for General-Domain dataset.}
\label{tab:general-result}
\end{table}

For the General-Domain dataset, we split the data into train (4,000 stories) and held-out test (200 stories) sets.
For each topic-specific set, we divided the hand-labeled data into a train (Train-HL) and held-out test, and created a second train set consisting of Train-HL and the data collected by bootstrapping (Train-HL-BS) as shown in Table~\ref{tab:train-test}. 
We automatically created a
question for every event occurring in the test data which resulted in 3,123 questions for General-Domain data, 
2,058 for the Camping and 2,533 questions for the
Storm topic.

\begin{table}[t]
\centering
\small
\begin{tabularx}{\columnwidth}{p{0.45in} X p{0.75in} p{0.45in}}
\toprule
{\bf Topic} & {\bf Model} & {\bf Train Dataset} & {\bf Accuracy} \\
\midrule
Camping &  Event-Unigram & Train-HL-BS & 0.507 \\
  Trip & Event-Bigram & Train-HL-BS & 0.510 \\
 & Event-SCP & Train-HL-BS & 0.508 \\
 &  Causal Potential & Train-HL & 0.631 \\
 &  Causal Potential & Train-HL-BS & 0.685 \\
\midrule
Storm & Event-Unigram & Train-HL-BS & 0.510 \\
 & Event-Bigram & Train-HL-BS & 0.523 \\
 & Event-SCP & Train-HL-BS & 0.516 \\
 & Causal Potential & Train-HL & 0.711 \\
 &  Causal Potential & Train-HL-BS & 0.887 \\
\bottomrule
\end{tabularx}
\caption{Automatic two-choice test results for Topic-Specific dataset.}
\label{tab:all-result}
\end{table}

For each dataset, we applied the baseline methods and
Causal Potential model on the train sets to learn contingent event pairs and
tested the pair collections on the questions generated from held-out test set.
We extracted about 418K contingent event pairs from General-Domain train set, 437K from Storm Train-HL-BS and 630K pairs from Camping Trip Train-HL-BS set using Causal Potential model.  We used our automatic test approach to evaluate these event pair collections.
The results for General-Domain and Topic-Specific datasets are shown in Table~\ref{tab:general-result} and Table~\ref{tab:all-result} respectively.

\begin{figure}[t!]
\small
\begin{tabularx}{\columnwidth}{p{0.03in} X}
\toprule
1 & go (nsubj:{\sc person}) $\rightarrow$ go (dobj:trail , prt:down)\\
2 & find (nsubj:{\sc person} , dobj:fellow)  $\rightarrow$ go (prt:back) \\
3 & see (nsubj:{\sc person} , dobj:gun)  $\rightarrow$ see (dobj:police)\\
4 & go (nsubj:{\sc person}) $\rightarrow$ go (nsubj:{\sc person} , dobj:rafting) \\
5 & come (nsubj:{\sc person}) $\rightarrow$ go (nsubj:{\sc person}) \\
6 &go (prt:out) $\rightarrow$ find (nsubj:{\sc person} , dobj:sconce) \\
7 & go (nsubj:{\sc person}) $\rightarrow$ see (dobj:window, prt:out) \\
8 & go (nsubj:{\sc person}) $\rightarrow$ walk (dobj:bit , prt:down) \\
\bottomrule
\end{tabularx}
\caption{Examples of event pairs with high CP scores extracted from General-Domain stories.}
\label{fig:general}
\end{figure}

The Causal Potential model trained on
Train-HL-BS dataset achieved accuracy of 0.685 on Camping Trip and 0.887
on Storm topic which is significantly stronger than all the baselines. Our
experiments indicate that having more training data collected by bootstrapping improves the accuracy of the model in predicting contingency relation between
events. Additionally, the Causal Potential results on Topic-Specific dataset is significantly stronger than General-Domain narratives indicating that using a topic-sorted dataset improves learning causal knowledge about events.  Fig.~\ref{fig:general} shows some examples of event pairs with high CP scores extracted from general-Domain set. In the following section we extract topic-indicative contingent event pairs and show that Topic-Specific data enables learning of finer-grained event knowledge that pertain to a particular theme.

\subsection{Topic-Indicative Contingent Event Pairs}
\label{sub:indic}
\label{sub:amt}

We identify contingent event pairs that are highly indicative of a
particular topic. We hypothesize that these event pairs serve as
building blocks of coherent event chains and narrative schema
since they encode contingency relation and correspond to a
specific theme. We evaluate the pairs on Amazon Mechanical Turk (AMT).

To identify event sequences that have a strong correlation to a topic
(topic-indicative pairs) we applied two filtering methods. First, we
selected the frequent pairs for each topic and removed the ones that
occur less than 5 times in the corpus.  Second, we used the indicative
event-patterns for each topic and extracted the pairs that at least
included one of these patterns. Indicative event-patterns are
automatically generated during the bootstrapping using AutoSlog-TS
and mapped to their
corresponding event representation as described in Sec.~\ref{sec:data}. 
Then we used the Causal Potential scores from our contingency model
for ranking the topic-indicative event pairs to
identify the highly contingent ones.  We sorted the pairs based on the
Causal Potential score and evaluated the top N pairs in this list. \\

\begin{table}[t]
\centering
\small
\begin{tabularx}{\columnwidth}{X | p{0.45in} | p{0.4in}}
\toprule
{\bf Label} & {\bf Camping} & {\bf Storm} \\
\midrule
Contingent \& Strongly Relevant & 44 \% & 33 \% \\
Contingent \& Somewhat Relevant & 8 \% & 20 \% \\
Contingent \& Not Relevant & 30 \% & 24 \% \\
\midrule
Total Contingent & 82 \% & 77 \%  \\
\bottomrule
\end{tabularx}
\caption{Results of evaluating indicative contingent event pairs on AMT.}
\label{tab:amt-result}
\end{table}

\begin{table*}
\small
\centering
\begin{tabularx}{6.3in}{X | p{2.9in}  p{2.4in}}
\toprule
{\bf Topic} & {\bf Label $\textgreater$ 2 : Contingent \& Strongly Topic-Relevant} & {\bf Label $\textless$ 1 : Not Contingent} \\
\midrule
Camping & person - pack up $\rightarrow$ person - go - home & person - pick up - cup $\rightarrow$ person - swim \\
Trip & person - wake up $\rightarrow$ person - pack up - backpack & pack up - tent $\rightarrow$ check out - video \\
 & person - head $\rightarrow$ hike up & person - play $\rightarrow$ person - pick up - sax \\
 & climb $\rightarrow$ person - find - rock & pack up - material $\rightarrow$ switch off - projector \\
 & person - pack up - car $\rightarrow$ head out & person - pick up - photo  $\rightarrow$ person - swim \\
\midrule
Storm & wind - blow - transformer $\rightarrow$ power - go out &  restore - community $\rightarrow$ hurricane - bend\\
 & tree - fall - eave $\rightarrow$ crush & boil $\rightarrow$ tree - fall - driveway \\
 & Ike - blow $\rightarrow$ knock down - limb & clean up - person $\rightarrow$ people - come out \\
 & air - push - person $\rightarrow$ person - fall out & blow - sign $\rightarrow$ person - sit \\
 & hit - location $\rightarrow$ evacuate - person  & person - rock - way $\rightarrow$ bottle - fall \\
\bottomrule
\end{tabularx}
\caption{Examples of event pairs evaluated on AMT.}
\label{tab:pairs}
\end{table*}

\noindent \textbf{Evaluations and Results.}
We evaluate the indicative contingent event pairs using human judgment on Amazon Mechanical Turk (AMT). Narrative schema consists of chains of events that are related in a coherent way and correspond to a common theme. Consequently, we evaluate the extracted pairs based on two main criteria:
\begin{itemize}
\item \textbf{Contingency:} Two events in the pair are likely to occur together in the given order and the second event is contingent upon the first one.
\item \textbf{Topic Relevance:} Both events strongly correspond to the specified topic.
\end{itemize} 

We have designed one task to assess both criteria since if an event pair is not contingent, it cannot be used in narrative schema for not satisfying the required coherence (even if it is topic-relevant). We asked the AMT annotators to rate each pair on a scale of 0-3 as follows:

\begin{quote}
\textbf{0:} The events are not contingent. \\
\textbf{1:} The events are contingent but not relevant to the specified topic. \\
\textbf{2:} The events are contingent and somewhat relevant to the specified topic. \\
\textbf{3:} The events are contingent and strongly relevant to the specified topic.
\end{quote}

To ensure that the Amazon Mechanical Turk annotations are reliable, we designed a \textit{Qualification Type} which requires the workers to pass a test before they can annotate our pairs. If the workers score 70\% or more on the test they will qualify to do the main task. For each topic we created a Qualification test consisting of 10 event pairs from that topic that were annotated by two experts. To make the events more readable for the annotators we used the following representation:

\begin{quotation}
Subject - Verb Particle - Direct Object
\end{quotation}

For example, {\small{\tt hike(subj:person, dobj:trail, prt:up)}} is mapped to {\small{\tt person - hike up - trail}}. For each topic we evaluated top $N=100$ event pairs and assigned 5 workers to rate each one. We generated a gold standard label for each pair by averaging over the scores assigned by the annotators and interpreted the average as follows: \\

\noindent \textbf{Label \textgreater 2:} Contingent \& strongly topic-relevant. \\
\noindent \textbf{Label = 2:} Contingent \& somewhat topic-relevant. \\
\noindent \textbf{1 $\leq$ Label $\textless$ 2:} Contingent \& not topic-relevant.\\
\noindent \textbf{Label $\textless$ 1:} Not contingent. \\

To assess the inter-annotator reliability we calculated kappa between each worker and the majority of the labels assigned to each pair. The average kappa was 0.73 which indicates substantial agreement.
The results in Table~\ref{tab:amt-result} show that 52\% of the Camping Trip and 53\% of the Storm pairs were labeled as contingent and topic-relevant by the annotators. The results also indicate that our model is capable of identifying event pairs with strong contingency relations: 82\% of the Camping Trip pairs and 77\% of the Storm pairs were marked as contingent by the workers.
Examples of the strongest and weakest pairs evaluated on Mechanical Turk are shown in Table~\ref{tab:pairs}. By comparison to Fig.~\ref{fig:general}, we can see that we can learn finer-grained type of events knowledge from topic-specific stories as compared to general-domain corpus.

\section{Discussion and Conclusions}
\label{sec:conclusion}
We learned fine-grained common-sense knowledge about contingent
relations between everyday events from personal stories written by ordinary people. 
We applied a semi-supervised bootstrapping approach using event-patterns to create topic-sorted sets of stories and evaluated our methods on a set of general-domain narratives as well
as two topic-specific datasets. 
We developed a new method for learning contingency relations between events that is tailored to the ``oral narrative" nature of the blog stories. Our evaluations indicate that a method that works well on the news genre does not generate coherent results on personal stories (comparison of Event-SCP baseline with Causal Potential).

We modeled the contingency (causal and conditional) relation between
the events from each dataset using Causal Potential and evaluated on the questions automatically generated from a held-out test set. 
The results show significant improvement over the Event-Unigram, Event-Bigram, and Event-SCP (Rel-grams method) baselines on Topic-Specific stories: 25\% improvement of accuracy on Camping Trip and 41\% on Storm topic compared to Bigram model.
In our future work, we plan to explore existing topic-modeling algorithms to create a broader set of topic-sorted corpora for learning contingent event knowledge.


Our experiments show that most of the fine-grained contingency relations we learn from narrative events are not found in existing narrative and event schema collections induced from the newswire datasets (Rel-grams).  
We also extracted indicative contingent event pairs from each topic and
evaluated them on Mechanical Turk. The evaluations show that 82\% of the relations between
events that we learn from topic-sorted stories are judged as contingent.  
We publicly release the extracted pairs
for each topic. In future work,
we plan to use the contingent event pairs as building blocks for
generating coherent event chains and narrative schema on several
different themes.


\bibliography{causal_pairs_acl16}
\bibliographystyle{acl2016}

\end{document}